\documentclass[10pt,twocolumn,letterpaper]{article}

\usepackage{cvpr}              

\usepackage{graphicx}
\usepackage{amsmath}
\usepackage{amssymb}
\usepackage{booktabs}
\usepackage{algorithm}
\usepackage{algpseudocode}
\usepackage{comment}
\usepackage[dvipsnames]{xcolor}
\usepackage{footnote}
\usepackage[font=small,skip=2pt]{caption}
\usepackage{multirow}

\usepackage[pagebackref,breaklinks,colorlinks]{hyperref}

\usepackage[capitalize]{cleveref}
\crefname{section}{Sec.}{Secs.}
\Crefname{section}{Section}{Sections}
\Crefname{table}{Table}{Tables}
\crefname{table}{Tab.}{Tabs.}

\def\cvprPaperID{*****} 
\def\confName{ECCV AROW Workshop}
\def\confYear{2022}

\begin{document}

\title{BARReL: Bottleneck Attention for Adversarial Robustness\\ in Vision-Based Reinforcement Learning}

\author{
  Eugene Bykovets\thanks{Equal contribution} \\
  D-INFK, ETH Zürich\\
  \texttt{eugene.bykovets@inf.ethz.ch} \\
  \and
  Yannick Metz$^\ast$\\\
  University of Konstanz\\
  \texttt{yannick.metz@uni-konstanz.de} \\
  \and
  Mennatallah El-Assady \\
  ETH AI Center, Zürich\\
  \texttt{melassady@ethz.ch} \\
  \and
  Daniel A. Keim \\
  University of Konstanz\\
  \texttt{keim@uni-konstanz.de} \\
  \and
  Joachim M. Buhmann \\
  D-INFK,   ETH Zürich\\
  \texttt{jbuhmann@inf.ethz.ch} \\
}

\maketitle
\begin{abstract}
Robustness to adversarial perturbations has been explored in many areas of computer vision. This robustness is particularly relevant in vision-based reinforcement learning, as the actions of autonomous agents might be safety-critic or impactful in the real world. We investigate the susceptibility of vision-based reinforcement learning agents to gradient-based adversarial attacks and evaluate a potential defense. We observe that Bottleneck Attention Modules (BAM) included in CNN architectures can act as potential tools to increase robustness against adversarial attacks. We show how learned attention maps can be used to recover activations of a convolutional layer by restricting the spatial activations to salient regions. Across a number of RL environments, BAM-enhanced architectures show increased robustness during inference. Finally, we discuss potential future research directions.
\vspace{-1em}
\end{abstract}
\section{Introduction}
\vspace{-0.3em}
\noindent Visual-based reinforcement learning (RL) is a sub-field of reinforcement learning research that uses an image as input for the decision-making process. Prominent examples include video games like Atari \cite{Mnih2015, bellemare13arcade} or robotics applications. Learning from raw pixels is a promising approach because it is generally applicable and requires little feature engineering, but it can be challenging in practice. While RL has profited from advances in deep learning architectures for computer vision, it has inherited its susceptibility to adversarial attacks \cite{lin2017tactics, qiaoben2021strategicallytimed}. In fact, trained RL agents are often very sensitive to the model to the quality of the visual input, i.e., they are easily corrupted due to environmental/equipment factors, like different lighting conditions, shadowing, camera quality/damage, or by adversarial intentions of the third parties, like adversarial attacks. The robustness of such systems is crucial for potential future real-world use, especially in safety-critical applications. In this work, we (1) propose a conceptually simple, general, yet effective inference-time method for defense against adversarial attacks. (2) In initial experiments, we demonstrate the effectiveness of the approach. Finally, we (3) discuss potential future research directions and extensions.
\begin{figure*}
    \centering
  \includegraphics[width=0.9\textwidth]{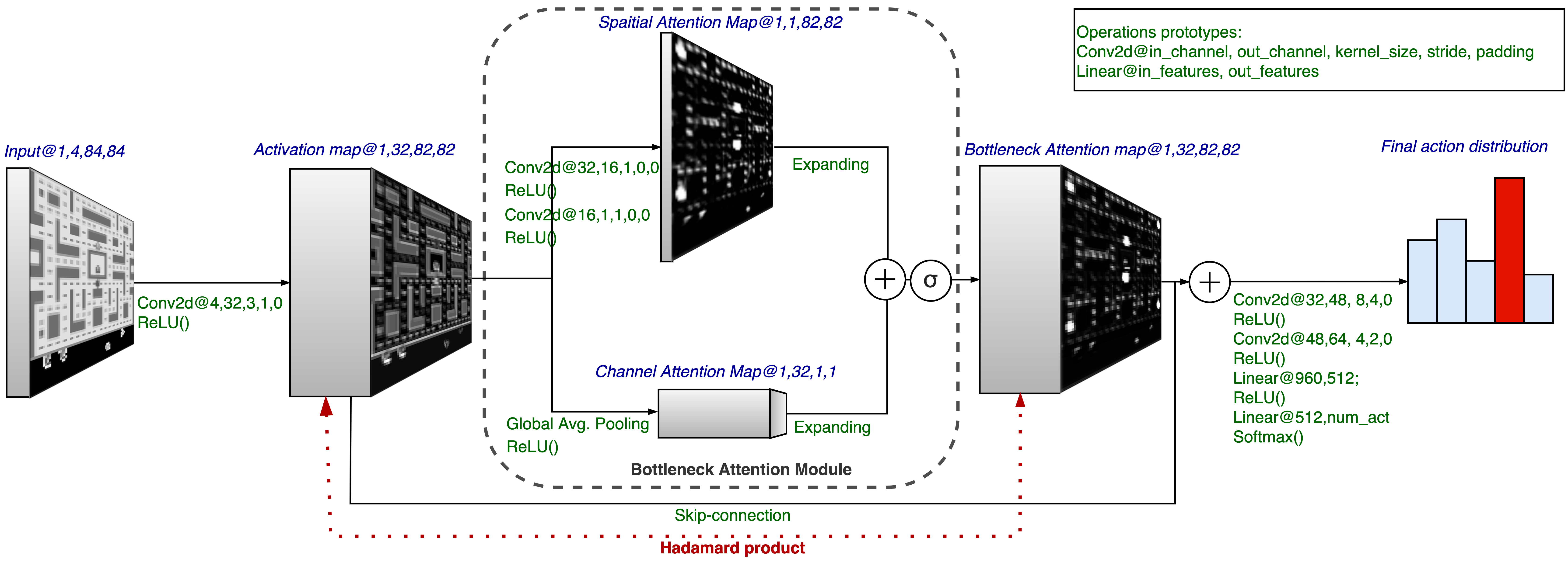}
  \caption{We specify all the tensors on top of picture of the tensor with \textcolor{Blue}{\textit{blue italic font}} and operations with \textcolor{OliveGreen}{green font} providing legend in top-right corner. The architecture of the \textcolor{gray}{\textbf{encircled Bottleneck Attention Module}} can be found in Park et al.\cite{bam}. The \textcolor{BrickRed}{dotted red line} indicates the elements used in the "recovery" mechanism based on attribution maps.\vspace{-0.5em}}
  \label{fig:overview_scheme}
\end{figure*}
\vspace{-1em}
\section{Related work}
\vspace{-0.3em}
\noindent\textbf{Adversarial attacks in Reinforcement Learning}
There are a number of previous works that tackle the problem of adversarial attacks on reinforcement learning agents. In RL, the goal of adversarial attacks is to deteriorate the policy's performance. Similar to the supervised settings, attacks can be classified into \textit{white-} and \textit{black-box} attacks \cite{ilahi2021challenges}. Beyond attacks found in supervised settings, which are comparable to attacking the state/observation space of an agent  \cite{qiaoben2021strategicallytimed, chen2021attention_adversarial}, there exist additional attacks such as poisoning the environment dynamics or reward function \cite{ilahi2021challenges}. In this work, we focus on white-box state space attacks, i.e., attacks on the input observations in which the attacker has access to the underlying model.\\
\noindent\textbf{Attention methods in Computer Vision}
Different regions of an image are not equally important for predictions. Identifying the most important part can be done via attention mechanisms. There are various approaches to model attention in computer vision \cite{guo2022attention}. In this work, we utilize \textit{Bottleneck Attention Modules} (BAM) \cite{bam} which applies both spatial and channel-wise attention. Following common practice, we utilize \textit{frame stacking} of a small sequence of proceeding game frames, which means that here channel-wise attention enables temporal attention.\\
\noindent\textbf{Defenses against adversarial attack}
Existing defense strategies include methods like adversarial training and game-theoretic approaches to defend against inference-time attacks or robust learning against training-time attacks \cite{ilahi2021challenges}. In this work, we combine adversarial training with a novel defense strategy to defend against inference-time attacks.

\section{Method}
\vspace{-0.5em}
\noindent For our approach, we draw inspiration from selective attention \cite{Tang2020}: only a part of visual input is necessary for decision making, i.e., a significant part of an input image contains non-relevant information. Gradient-based adversarial attacks modify both relevant and non-relevant parts of visual input. The entirety of the perturbation can corrupt the predicted output action distribution of an RL agent and consequently cause potentially erroneous behavior. Thus, our goal is to mitigate the effect of these attacks by removing non-relevant parts of the input, which can dramatically reduce the magnitude of the perturbation. We utilize \textit{Bottleneck Attention Module} \cite{bam}, which has been introduced as a simple and non-invasive extension to existing CNN architectures, to learn attention maps of the most relevant spatial and temporal features for vision-based RL. We use learned attention maps to \textit{"recover"} the output of a convolutional layer perturbed by an adversarial attack. In effect, we use the attention maps to only retain relevant parts of the input.

\begin{figure*}[h!]
  \centering
  \includegraphics[width=0.95\textwidth]{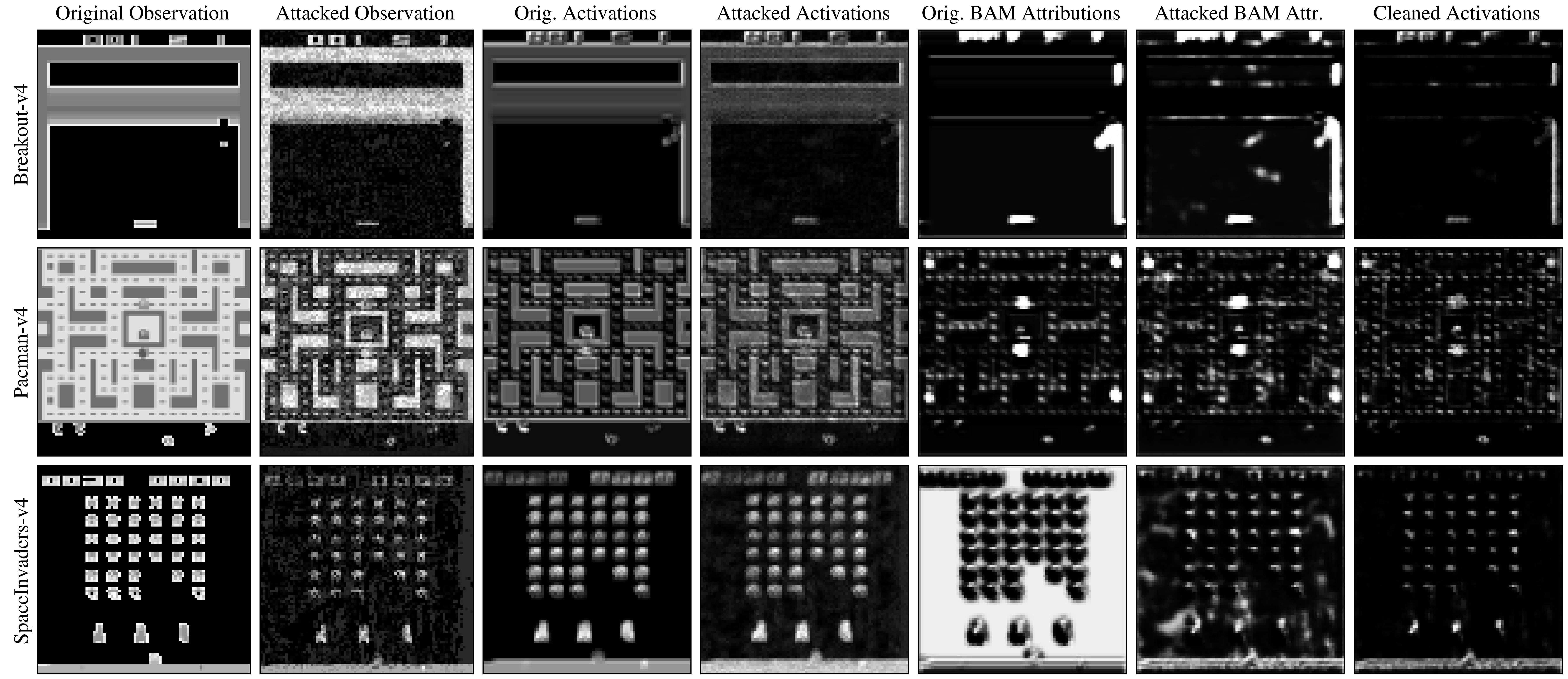}
  \caption{Qualitative sampled results of the proposed activation recovery process. All states were attacked with Linear PGD-method \cite{madry2018towards} with an $\epsilon$ of $0.1$. For the observations, only the last channel is displayed (corresponding to the most frequent time step). All activations and attention are averaged over channels. While the attacks are very noticeable both in the activations and attention maps, the cleaned map removes a significant of the perturbations while retaining all necessary information for decision making.\vspace{-0.5em}}
  \label{fig:visual_resuls}
\end{figure*}

\subsection{Implementation Details}
\label{subsec:training_details}
\vspace{-0.3em}
\noindent \textbf{Training} We use Proximal Policy Optimization \cite{schulman2017proximal}, with the implementation and default training hyper-parameters from the \texttt{StableBaseline3} library\cite{Raffin2021}. We train all models on three Atari environments, which are the part of a well-established benchmark used to compare RL methods~\cite{bellemare13arcade}, namely: \texttt{Breakout-v4}, \texttt{SpaceInvaders-v4}, \texttt{MsPacman-v4}. We apply the common pre-processing routing for observation; namely, transformation to greyscale, resizing, a frame-skipping of 4 frames, as well as stacking of the 4 last frames to encode temporal dynamics. As a baseline architecture (\texttt{Nature-CNN}), we use a 3-layer CNN, followed by two fully-connected layers and a softmax that outputs a categorical probability distribution over a set of possible actions similar to \cite{Mnih2015}. We extend this architecture by embedding a \textit{BAM}-layer \cite{bam} between the first and second convolutional layer (\texttt{BAM-CNN}). These architectures contain 2.246.416 parameters (\texttt{Nature-CNN}) and 2.248.409 parameters (\texttt{BAM-CNN}), respectively. Figure~\ref{fig:overview_scheme} shows the used architecture. To have a performance baseline, we trained both Nature-CNN and BAM-CNN architectures on the default environments (denoted without any additional suffix in Tab.~\ref{tab:reward_results}). We utilize adversarial training by applying Projected Gradient Descend (PGD) attacks \cite{madry2018towards} using \texttt{Foolbox} framework \cite{rauber2017foolbox} with an $\epsilon=0.1$ and get two additional training regimes: \texttt{Nature-CNN-Adv.} and \texttt{BAM-CNN-Adv.}. In RL, the quality of training is directly dependent on the learned policy, thus applying attacks too aggressively, e.g., at every step, might stop an agent from learning altogether. We found that attacking only every 10th frame had no significant impact on training performance see Tab.~\ref{tab:reward_results}), but in turn only gave moderate robustness to adversarial attacks (Tab.~\ref{tab:defence_results} shows that adversarial attacks are successful with approx. $100\%$ probability). However, our experiments showed that adversarial training seems to be necessary to learn appropriate attention maps. When trained on the default environment, the learned attention maps were generally not suited to support the recovery of activations.\\
\noindent\textbf{Inference} Performance of the RL agents at the inference stage with respect to cumulative reward, averaged over 10 episodes, for all of the environments is presented in Tab. \ref{tab:reward_results} that empirically confirms that both: (1) introducing of BAM layer to architecture and (2) incorporation of adversarial samples during training do not lead to dropping of the performance.
\vspace{-0.3em}
\subsection{Recovery algorithm}
\vspace{-0.3em}
 \noindent We want to describe the algorithm we use to 
"recover" the intermediate activations with the learned attention maps. $\psi_\theta$ denotes the \texttt{BAM-CNN-Adv.} trained model (see Sec.~\ref{subsec:training_details}), parameterized by $\theta$ with $N$ layers with a BAM layer placed at index $l$. 
\begin{algorithm}[h]
\caption{Adversarial Attack recovery}\label{alg:cap}
\begin{algorithmic}

\State \textbf{Input:} $\psi_\theta$
\State \textbf{Output:} $\tilde{\pi}(a|s_A)$ that is close to $\pi^{*}(a|s)$
\State $f^{(l-1)} \gets \psi^{[1:l-1]}_{\theta}(s_A)$ \Comment{For notation clarification see \footnotemark}.
\State $f^{BAM} \gets \psi^{[1:l]}_{\theta}(s_A)$ 
\State $f^{(l-1)}_{rec.} \gets f^{(l-1)}\odot f^{BAM}$
\State $\tilde{\pi}(a|s_A) = \psi^{[l:N]}_{\theta}(f^{(l-1)}_{rec.})$
\end{algorithmic}
\label{alg:defence}
\end{algorithm}
Adversarial examples $s_A=\mathcal{A}(\psi_{\theta},s,\epsilon)$ for original states $s$ are created by an attack algorithm $\mathcal{A}$, where $\epsilon$ controlling the severity of the perturbation. The goal of the proposed algorithm is to recover a policy $\tilde{\pi}(a|s_A)$ that is close to the original policy $\pi^{*}(a|s)$ induced by $\psi_{\theta}$. The policies' similarity metrics are discussed in Sec. \ref{sec:results}. We present the pseudo-code of the inference-time defense algorithm in Alg. \ref{alg:defence}. Our analysis shows that this particular reinforcement learning scenario produces highly polarized attention maps (i.e., very high contrast between high- and low-attention areas). Therefore, the learned BAM maps have a strong notion of what is important for decision-making. 
As a result, applying element-wise (Hadamard) multiplication allows us to "prune" part of the input that is not significant for the decision-making but still affected by adversarial perturbations.
\footnotetext{{$\psi^{[i:j]}_{\theta}(x)$ notation means slicing $\psi_{\theta}$ from layer with index $i$ to layer with index $j$ and apply to tensor $x$.}}
\section{Results}
\label{sec:results}
\vspace{-1em}
\begin{table}[h!]
\centering
\resizebox{\columnwidth}{!}{%
\begin{tabular}{||l|l|l|l||}
\hline\hline
                & Breakout-v4    & MsPacman-v4        & SpaceInvaders-v4  \\ \hline\hline
Existing Best   & $1.5_{\pm 0.9}$  & $504.0_{\pm 158.25}$ & $166.5_{\pm 74.8}$  \\ \hline
BAM-CNN-Adv.+Recovery & $6.3_{\pm 6.84}$ & $687.0_{\pm 242.0}$  & $170.0_{\pm 107.5}$ \\ \hline
Difference      & $\times 4.2$     & $\times 1.36$        & $\times 1.02$      \\ \hline\hline
\end{tabular}%
}
\caption{Comparison between Baseline and BamCNN+Recovery: We compare average rewards of best baseline method (see Tab.~\ref{tab:reward_results}, e.g.\texttt{Nature-CNN-Adv.}) compared to our attention-map based defense method (\texttt{BAM-CNN-Adv.+Recovery}).\vspace{-0.5em}}
\label{tab:our_results}
\end{table}
\begin{table*}[h]
\resizebox{\textwidth}{!}{%
\begin{tabular}{||l||llll||llll||llll||}
\hline\hline
& \multicolumn{4}{l||}{\centering Breakout-v4} & \multicolumn{4}{l||}{MsPacman-v4} & \multicolumn{4}{l||}{SpaceInvaders-v4} \\ \cline{2-13} 
 & \multicolumn{1}{l|}{Nat.-CNN} & \multicolumn{1}{l|}{BAM-CNN} & \multicolumn{1}{l|}{Nat.-CNN-Adv.} & \multicolumn{1}{l||}{BAM-CNN-Adv.} & \multicolumn{1}{l|}{Nat.-CNN} & \multicolumn{1}{l|}{BAM-CNN} & \multicolumn{1}{l|}{Nat.-CNN-Adv.} & \multicolumn{1}{l||}{BAM-CNN-Adv.} & \multicolumn{1}{l|}{Nat.-CNN} & \multicolumn{1}{l|}{BAM-CNN} & \multicolumn{1}{l|}{Nat.-CNN-Adv.} & \multicolumn{1}{l||}{BAM-CNN-Adv.} \\ \hline \hline
 Default env. & \multicolumn{1}{l|}{$166.2_{\pm 105.3}$} & \multicolumn{1}{l|}{$59.0_{\pm 15.5}$} & \multicolumn{1}{l|}{$98.4_{\pm 82.2}$} & \multicolumn{1}{l||}{$96.8_{\pm 38.1}$} & \multicolumn{1}{l|}{$1736.0_{\pm 98.5}$} & \multicolumn{1}{l|}{$2147.0_{\pm 59.7}$} & \multicolumn{1}{l|}{$1802.0_{\pm 230.4}$} & \multicolumn{1}{l||}{$1826.0_{\pm 47.4}$} & \multicolumn{1}{l|}{$700.5_{\pm 217.9}$} & \multicolumn{1}{l|}{$815.0_{\pm 217.9}$} & \multicolumn{1}{l|}{$737.0_{\pm 350.2}$} & \multicolumn{1}{l||}{$734.5_{\pm 276.8}$} \\ \hline
 Attacked env. & \multicolumn{1}{l|}{$0.4_{\pm 1.2}$} & \multicolumn{1}{l|}{$0.5_{\pm 1.5}$} & \multicolumn{1}{l|}{$0.2_{\pm 0.6}$} & \multicolumn{1}{l||}{$1.5_{\pm 0.9}$} & \multicolumn{1}{l|}{$504.0_{\pm 158.25}$} & \multicolumn{1}{l|}{$70.0_{\pm 0.0}$} & \multicolumn{1}{l|}{$297.0_{\pm 150.4}$} & \multicolumn{1}{l||}{$253.0_{\pm 131.4}$} & \multicolumn{1}{l|}{$29.5_{\pm 24.5}$} & \multicolumn{1}{l|}{$33.0_{\pm 14.3}$} & \multicolumn{1}{l|}{$166.5_{\pm 74.8}$} & \multicolumn{1}{l||}{$42.0_{\pm 21.9}$} \\
\hline \hline
\end{tabular}%
}
\caption{Cumulative reward results of baselines: \texttt{Nature-CNN}, \texttt{BAM-CNN}, \texttt{Nature-CNN-Adv.}, \texttt{BAM-CNN-Adv.} for \texttt{Breakout-v4}, \texttt{SpaceInvaders-v4}, \texttt{MsPacman-v4} averaged over 10 episodes with standard deviations. For the adv. attacks, each frame is attacked with an $\epsilon = 0.05$.\vspace{-0.3em}. We contrast these scores with the reward achieved by our proposed method (see Tab.~\ref{tab:our_results}).}
\label{tab:reward_results}
\end{table*}

\begin{table*}[h!]
\resizebox{\textwidth}{!}{%
\begin{tabular}{||l||r|r|r|r||r|r|r|r||r|r|r|r||}
\hline \hline
 & \multicolumn{4}{l||}{Breakout-v4} & \multicolumn{4}{l||}{MsPacman-v4} & \multicolumn{4}{l||}{SpaceInvaders-v4} \\ \cline{2-13} 
 & \multicolumn{1}{l|}{$\epsilon=0.01$} & \multicolumn{1}{l|}{$\epsilon=0.05$} & \multicolumn{1}{l|}{$\epsilon=0.1$} & \multicolumn{1}{l||}{$\epsilon=0.5$} & \multicolumn{1}{l|}{$\epsilon=0.01$} & \multicolumn{1}{l|}{$\epsilon=0.05$} & \multicolumn{1}{l|}{$\epsilon=0.1$} & \multicolumn{1}{l||}{$\epsilon=0.5$} & \multicolumn{1}{l|}{$\epsilon=0.01$} & \multicolumn{1}{l|}{$\epsilon=0.05$} & \multicolumn{1}{l|}{$\epsilon=0.1$} & \multicolumn{1}{l||}{$\epsilon=0.5$} \\ \hline \hline
   \text{$\%$ Successful Attacks} & $66.22$ &  $93.42$ &  $95.76$ &    $84.54$ &   $74.0$ &   $100.0$ &  $100.0$ & $100.0$ &      $94.0$ & $100.0$ &   $100.0$ &   $100.0$ \\ \hline
  \text{Reversed-TOP-1} & $57.78$ & $35.51$ & $32.73$ &  $15.45$ &   $30.73$ & $22.44$ & $19.74$ & $0.0$ &       $13.67$ &  $2.6$ &  $0.045$ & $0.0$  \\ \hline
  \text{Reversed-TOP-2} & $89.53$ & $76.59$ & $66.36$ &  $66.10$ &   $50.62$ &    $44.15$ &  $34.627$ & $5.6$ &       $40.15$ &  $17.77$ & $7.2$ & $0.24$ \\ \hline
  \text{Reversed-ANY} & $65.01$ & $49.75$ & $51.41$ & $15.45$ &  $83.25$ &  $70.39$ & $70.54$ &  $0.35$ &       $41.24$ & $35.59$ & $31.25$ &  $0.24$ \\ \hline \hline
\end{tabular}}

\caption{Attack recovery results with different $\epsilon$ values for \texttt{Breakout-v4}, \texttt{SpaceInvaders-v4}, \texttt{MsPacman-v4}, averaged over 10 games each. The used architecture was trained on partly attacked data during training (with every 10th frame perturbed), which only resulted in marginal adversarial robustness. This adversarial training by itself is not able to effectively defend against adversarial attacks: As is visible in the first line, adversarial attacks manage to switch the selected action in most cases, even for small $\epsilon$. Applying the presented recovery technique can decrease the impact of attacks up to a certain level.\vspace{-0.3em}}
\label{tab:defence_results}
\end{table*}
\vspace{-0.5em}
\noindent Tab.~\ref{tab:our_results} and Tab.~\ref{tab:reward_results} summarize the reward performance of different models. Tab.~\ref{tab:reward_results} e.g. shows that adversarial training can improve performance on non-attacked (default) environments. We hypothesize, that the limited adversarial attacks can act as a type of regularization/data augmentation. However, as mentioned above, adversarial training generally does not lead to huge robustness gains, i.e. the reward for attacked environments is noticeably worse in any case. Tab.~\ref{tab:our_results} summarizes the performance gains of our method against the strongest baseline method. Besides reward performance, we evaluate our method by measuring the following metrics: (a) The percentage of fully reversed attacks (\texttt{Reversed-TOP-1}), for which the action predicted based on the cleaned activations matches the original action based on the non-attacked state. (b) The percentage of partially reversed attack (\texttt{Reversed-TOP-2}) when the recovery leads to the predicted action matching either the first or second choice of the original action. (c) The percentage of partially reversed attack (\texttt{Reversed-ANY}) when the recovered action is different from the attacked action (or already the original action assigned maximum probability). We also report the (d) percentage of successful PGD attacks (\texttt{\% Successful Attacks}), i.e., how often the attack is able to switch the maximum probability action. The results for different $\epsilon$ across different environments can be found in Tab.~\ref{tab:defence_results}. Figure~\ref{fig:visual_resuls}
shows qualitative visual results of the "recovery" technique. The adversarial attack targets the entire frame, which is clearly visible in the activations. The recovered activation shows how a significant part of the applied perturbations is removed from the input image. As a result, the output action distribution after the "recovery" procedure is closer to the original action distribution (see Tab.~\ref{tab:defence_results}). For example, in \texttt{Breakout-v4} for moderate $\epsilon = \{0.01, 0.05\}$, the original action is restored with $57.78\%$ and $35.51\%$ probability, respectively (out of 4 total actions). For \texttt{MsPacman-v4}, for approx. $30\%$ of states the original action is restored (out of 9 possible actions). This recovery is noticeable in the final performance of the trained agents attacked with adversarial observations. We see improvements that scale with the success rate of the defense. 
\vspace{-1em}
\section{Discussion \& Future work}
\vspace{-0.3em}
\noindent 
A great strength of the presented method is it's simplicity and low application cost. The additional BAM-layer and clean-up process during inference only adds marginal complexity in terms of model parameters and inference time. BAM itself is suitable for any type of CNN architecture, but we might explore other types of attention (e.g., self-attention) for other architectures like vision transformers \cite{Dosovitskiy2020}. Another potential benefit is the use of selective attention to increase robustness against other types of distractors like background images. While the initial results showcase the potential of the method, the performance of agents is still worse in the attacked environments. It is, therefore, necessary to explore how to improve the method further. Another shortcoming is the fact, that we applied the presented defense in a grey-box setting, in which the attacker is not aware of the inference procedure. To further investigate robustness, we should explore adaptive attack procedures \cite{tramer2020} with the attacker having full knowledge of the attack. Furthermore, a comparison to other defense methods \cite{ilahi2021challenges} is warranted.

Besides the aforementioned improved evaluation, we plan to extend our method in follow-up work:
(a) Investigating its utility for core computer vision problems and different backbones that do not share characteristics of reinforcement learning and the investigated environments. (b) Investigate different "recovery" procedures on feature maps, e.g., using different operations besides element-wise multiplication. (c) Use multiple BAM attention layers in a network and apply the clean-up- procedure multiple times, which might increase robustness, (d) Investigate the possibility of application of fewer amount of attacks during training. (e) Explore different types of adversarial attack methods. (f) Extend the current method to vector-based RL and continuous action spaces. 
\vspace{-1em}
\section{Conclusion}
\vspace{-0.3em}
\noindent In this paper, we presented a simple adversarial defense method for deep RL based on \textit{Bottleneck Attention}. This type of selective attention could be a general tool to restrict the effect on non-targeted adversarial perturbations. We presented initial promising results that demonstrate the potential of the method to recover activations perturbed by adversarial attacks. In future work, we plan to apply a more thorough evaluation and study of potential improvements.



{\small
\bibliographystyle{ieee_fullname}
\bibliography{main}
}

\end{document}